\documentclass[runningheads,a4paper]{llncs}
\usepackage{float}
\usepackage{amsmath}
\usepackage{graphicx}

\usepackage{amssymb}
\usepackage{graphicx}
\usepackage{url}
\urldef{\mailsa}\path|dot@tx.technion.ac.il|
\urldef{\mailsb}\path|claudio.gentile@uninsubria.it|
\urldef{\mailsc}\path|shie@ee.technion.ac.il|

\usepackage{algorithmic}
\usepackage{algorithm}

\newcommand{\bx}{\boldsymbol{x}}
\newcommand{\bu}{\boldsymbol{u}}
\newcommand{\bw}{\boldsymbol{w}}

\newcommand{\figscale}[2]{\includegraphics[scale=#2,clip=false]{#1}}

\begin{document}

\mainmatter  

\title{Bandits with an Edge}

\titlerunning{Bandits with an edge}

%
%
\author{Dotan Di Castro$^1$
\and Claudio Gentile$^2$\and Shie Mannor$^1$}
\authorrunning{Di Castro, Gentile, Mannor}

\institute{$^1$Technion, Israel Institute of Technology, Haifa, Israel\\
$^2$Universita' dell'Insubria, Varese, Italy\\
\mailsa, \mailsb, \mailsc
}

%
%

\toctitle{Lecture Notes in Computer Science}
\tocauthor{Authors' Instructions}
\maketitle

\begin{abstract}
We consider a bandit problem over a graph where the rewards are not directly observed.
Instead, the decision maker can compare two nodes and receive (stochastic) information pertaining to the
difference in their value. The graph structure describes the set of possible comparisons.
Consequently, comparing between two nodes that are relatively far requires estimating the difference between every pair of nodes on the path between them.
We analyze this problem from the perspective of sample complexity: How many queries are needed to find an approximately optimal node with probability more than $1-\delta$ in the PAC setup?
We show that the topology of the graph plays a crucial in defining the sample complexity: graphs with a low diameter have a much better sample complexity.
\end{abstract}

\section{Introduction}
%
We consider a graph where every edge can be sampled. When sampling an edge, the decision maker
obtains a signal that is related to the value of the nodes defining the edge. The objective of the decision maker
is to locate the node with the highest value. Since there is no possibility to sample the value of the nodes directly, the decision maker has to infer which is the best node by considering the differences between the nodes.

As a motivation, consider the setup where a user interacts with a webpage. In the webpage, several links or ads can be presented, and the response of the user is to click one or none of them. Essentially, in this setup we query the user to compare between the different alternatives. The response of the user is comparative: a preference of one alternative to the other will be reflected in a higher probability of choosing the alternative. It is much less likely to obtain direct feedback from a user, asking her to provide an evaluation of the worth of the selected alternative. In such a setup, not all pairs of alternatives can be directly compared or, even if so, there might be constraints on the number of times a pair of ads can be presented to a user. For example, in the context of ads it is reasonable to require that ads for similar items will not appear in the same page (e.g., two competing brands of luxury cars will not appear on the same page).
In these contexts, a click on a particular link cannot be seen as an absolute relevance judgement~(e.g., \cite{j02}), but rather as a relative preference. Moreover, feedback can be noisy and/or inconsistent, hence aggregating the choices into a coherent picture may be a non-trivial task. Finally, in such contexts  pairwise comparisons occur more frequently than multiple comparisons, and are also
more natural from a cognitive point of view~(e.g.,\cite{t27}).

We model this learning scenario as bandits on graphs where the information that is obtained is differential.
We assume that there is an inherent and unknown value per node, and that the graph describes the allowed (pairwise)
comparisons.
That is, nodes $i$ and $j$ are connected by an edge if they can be compared by a single query. In this case, the query
returns a random variable whose distribution depends, in general, on the values of $i$ and $j$.
For the sake of simplicity, we assume that the observation of the edge between nodes $i$ and $j$ is a random variable
that depends only on the {\em difference} between the values of $i$ and $j$.
Since this assumption is restrictive in terms of applicability of the algorithms,
we also consider the more general setup where contextual information is observed before sampling the edges.
This is intended to model a more practical setting where, say, a web system has preliminary access to a set
of user profile features.

In this paper, our goal is to identify the node with the highest value, a problem that has been studied
extensively in the machine learning literature (e.g., \cite{evendar2006action,abm10}).
More formally, our objective is to find an approximately optimal node (i.e.,
a node whose value is at most
$\epsilon$ smaller than the highest value) with a given failure probability $\delta$ as quickly as
possible.
When contextual information is added, the goal becomes to progressively fasten the time
needed for identifying a good node for the given user at hand, as more and more users interact with
the system.

\noindent{\bf Related work. }
There are two common objectives in stochastic bandit problems: minimizing the regret
and identifying the ``best'' arm.
While both objectives are of interest, regret minimization seems particularly difficult in our setup.
In fact, a recent line of research related to our paper is the \emph{Dueling Bandits Problem}
of Yue et al. \cite{yue2011,yj11} (see also \cite{frpu94}).
In the dualing bandit setting, the learner
has at its disposal a {\em complete} graph of comparisons between pairs of nodes, and each edge
$(i,j)$ hosts an unknwon preference probability $p_{i,j}$ to be interpreted as the probability that
node $i$ will be preferred over node $j$. Further consistency assumptions (stochastic transitivity
and stochastic triangle inequality) are added. The complete graph assumption
allows the authors to define a well-founded notion of regret, and analyze a regret
minimization algorithm which is further enhanced in \cite{yj11}
where the consistency assumptions are relaxed.
Although at first look our paper seems to deal with the same setup, we highlight here the main
differences. First, the setups are different with respect to the topology handled.
In \cite{yue2011,yj11} the topology is always a complete graph which results in the possibility to directly
compare between every two nodes. In our work (as in real life) the topology is \emph{not} a complete
graph, resulting in a framework where a comparison of two nodes requires sampling all the edges between
the nodes. In the extreme case of a straight line we need to sample all the given edges in the graph
in order to compare the two nodes that are farthest apart.
Second, the objective of minimizing the regret is natural for a complete graph where it amounts to
comparing a choice of the best bandit repeatedly
with the actual pairs chosen. In a topology other than the complete graph this notion is less
clear since one has to restrict choices to edges that are available.
%
%
Finally, the algorithms in \cite{yue2011,yj11} are geared towards the elimination of
arms that are not optimal with high probability. In our setup one \emph{cannot} eliminate such nodes
and edges because it is crucial in comparing candidates for optimal nodes. Therefore, the resulting
algorithms and analyses are quite different.
On the other hand, constraining to a given set of allowed comparisons leads us to make less
general statistical assumptions than \cite{yue2011,yj11}, in that our algorithms are based on the
ability to reconstruct the reward difference on adjacent nodes by observing their connecting edge.


From a different perspective, the setup we consider is reminiscent of online learning with partial monitoring
\cite{LugosiMStoltz08Strategies}. In the partial monitoring setup, one usually does not observe the reward directly,
but rather a signal that is related (probabilistically) to the unobserved reward. However, as far we know,
the alternatives (called arms usually) in the partial monitoring setup are separate and there is no additional
structure: when sampling an arm a reward that is related to this arm alone is obtained but not observed.
Our work differs in imposing an additional structure, where the signal is derived from the structure of the
problem where the signal is always relative to adjacent nodes. This means that comparing two nodes that are not
adjacent requires sampling all the edges on a path between the two nodes. So that deciding which of two remote
nodes has higher value requires a high degree of certainty regarding all the comparisons on the path between them.


Another research area which is somewhat related to this paper is learning to rank via
comparisons
(a very partial list of references includes~\cite{crs99,j02,dms04,bsr05,jr05,cqltl07,hfcb08,xlwzl08,Liu09}).
Roughly speaking, in this problem we have a collection of training instances to be associated with a finite set of
possible alternatives or classes (the graph nodes in our setting). Every training example is assigned a set of
(possibly noisy or inconsistent) pairwise (or groupwise) preferences between the classes.
The goal is to learn a function
that maps a new training example to a total order (or ranking) of the classes. We emphasize that
the goal in this paper
is different in that we work in the bandit setup with a given structure for the comparisons and, in addition,
we are just aiming at identifying the (approximately) best class, rather than ranking them all.

\noindent{\bf Content of the paper. }
The rest of the paper is organized as follows. We start from the formal model in Section \ref{sec:model}.
We analyze the basic linear setup, where each node is comparable to at most two nodes in
Section~\ref{sec:Linear-Topology}.
We then move to the tree setup and analyze it in Section \ref{s:tree}.
The general setup of a network is treated in Section \ref{s:net}.
Some experiments are then presented in Section \ref{s:sims} to elucidate the theoretical findings in
previous sections.
In Section \ref{s:extensions} we discuss the more general setting with contextual information.
We close with some directions for future research.

\section{Model and Preliminaries}\label{sec:model}
In this section we describe the classical Multi-Armed Bandit (MAB)
setup, describe the Graphical Bandit (GB) setup, state/recall two
concentration bounds for sequences of random variables, and review a few terms from
graph theory.

\subsection{The Multi-Armed Bandit Problem}
The MAB model \cite{lai1985asymptotically} is comprised of a set of \emph{arms} $A\triangleq\left\{ 1,\ldots,n\right\} $.
When sampling arm $i\in A$, a \emph{reward} which is a random variable
$R_{i}$, is provided.
Let $r_{i}=\mathbb{E}\left[R_{i}\right]$. The goal in the MAB
setup is to find the arm with the highest expected reward, denoted
by $r^{*}$, where we term this arm's reward the \emph{optimal
reward}. An arm whose expected reward is strictly less than $r^{*}$
is called a \emph{non-best arm}. An arm $i$ is called an $\epsilon$-optimal
arm if its expected reward is at most $\epsilon$ from the optimal
reward, i.e., $\mathbb{E}\left[R_{i}\right]\ge r^{*}-\epsilon$. In
some cases, the goal in the MAB setup is to find an $\epsilon$-optimal arm.

A typical algorithm for the MAB problem does the following. At each
time step $t$ it samples an arm $i_{t}$ and receives a reward $R_{i_t}$.
When making its selection,
the algorithm may depend on the history (i.e., the actions and rewards)
up to time $t-1$. Eventually the algorithm must commit to a single
arm and select it. Next we define the desired properties of such
an algorithm.
\begin{definition}
(PAC-MAB) An algorithm is an $\left(\epsilon,\delta\right)$-probably
approximately correct (or $\left(\epsilon,\delta\right)$-PAC) algorithm
for the MAB problem with sample complexity $T$, if it terminates
and outputs an $\epsilon$-optimal arm with probability at least $1-\delta$,
and the number of times it samples arms before
termination is bounded by $T$.
\end{definition}
In the case of standard MAB problems there is no structure defined over
the arms. In the next section we describe the setup of our work where such a
structure exists.

\subsection{The Graphical Bandit Problem}
Suppose that we have an undirected and connected graph $G = (V,E)$ with nodes
$V=\left\{ 1,\ldots,n\right\}$ and edges $E$. The nodes are associated with reward
values $r_{1},\ldots,r_{n}$, respectively, that are unknown to us. We denote
the node with highest value by $i^{*}$ and, as before, $r^* = r_{i^*}$. Define
$u\triangleq \min_{j\ne i^*} r_{i^*}-r_{j}$ to be the difference between the node
with the highest value and the node with the second highest value. We call $u$ the reward
{\em gap}, and interpret it as a measure for how
easy is to discriminate between the two best nodes in the network.
As expected, the gap $u$ has a significant influence on the sample complexity bounds
(provided the accuracy parameter $\epsilon$ is not large).
We say that nodes $i$ and $j$ are \emph{neighbors}
if there is an edge in $E$ connecting them (and denote the edge random variable by $E^{ij}$).
This edge value is a random variable whose distribution is determined by
the nodes it is connecting, i.e.,  $(i,j)$'s statistics are determined by $r_{i}$
and $r_{j}$. In this work, we assume that\footnote{
Notice that, although the graph is undirected, we view edge $(i,j)$ as a directed
edge from $i$ to $j$. It is understood that $E^{ji} = -E^{ij}$.
}
$\mathbb{E}\left[E^{ij}\right]= r_{j}-r_{i}$.
Also, for the sake of concreteness, we assume the edge values are bounded in $[-1,1]$.

In this model, we can only sample the graph edges $E^{ij}$ that provide independent
realizations of the node differences.
For instance, we may interpret $E^{ij} = +1$ if the feedback we receive says that
item $j$ is preferred over item $i$, $E^{ij} = -1$ if $i$ is preferred over $j$, and
$E^{ij} = 0$ if no feedback is received. Then the reward difference $r_j - r_i$
becomes equal to the difference between the probability of preferring
$j$ over $i$ and the probability of preferring $i$ over $j$.
Let us denote the realizations of $E^{ij}$ by $E_{t}^{ij}$
where the subscript $t$ denotes time.
Our goal is to find an $\epsilon$-optimal
node, i.e., a node $i$ whose reward satisfies $r_{i}\ge r^{*}-\epsilon$.

Whereas neighboring nodes can be directly compared by sampling its connecting
edge, if the nodes are far apart, a comparison between the two can only be done
indirectly, by following a path connecting them.
We denote a \emph{path} between node $i$ and node $j$ by $\pi_{ij}$.
Observe that there can be several paths in $G$ connecting $i$ to $j$.
For a given path $\pi$ from $i$ to $j$, we define the
\emph{composed edge} value $E_{\pi}^{ij}$ by $E_{\pi}^{ij} = \sum_{(k,l)\in\pi_{ij}} E^{kl}$,
with $E_{\pi}^{ii} = 0$.
By telescoping, the average value of a composed edge $E^{ij}$ only depends
on its endpoints, i.e.,
\begin{equation}\label{eq:depends_edge}
\mathbb{E}\left[E_{\pi}^{ij}\right]=\sum_{(k,l)\in\pi_{ij}}\mathbb{E}\left[E^{kl}\right]
=\sum_{(k,l)\in\pi_{ij}}\left(r_{l}-r_{k}\right) = r_{j}-r_{i},
\end{equation}
independent of $\pi$.
Similarly, define $E_{t}^{ij}$ to be the time-$t$ realization
of the composed edge random variable $E_{\pi}^{ij}$ when we pull once
all the edges along the path $\pi$ joining $i$ to $j$. A schematic illustration of the the GB setup is presented in Figure 1.

\vspace{0.0in}
\begin{figure}[htp]
\centering
\includegraphics[scale=0.6]{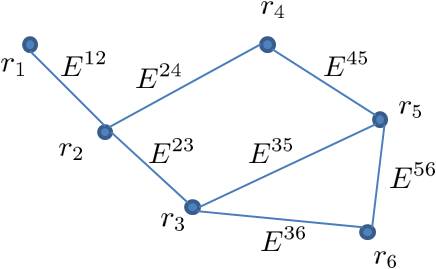}
\caption{Schematic illustration of the the GB setup for 6 nodes}\label{fig:setup}
\end{figure}

The algorithms we present in the next sections hinge on constructing reliable estimates
of edge reward differences, and then combining them into a suitable node selection procedure.
This procedure heavily depends on the graph topology.
In a tree-like (i.e., acyclic) structure no inconsistencies can arise due to the noise
in the edge estimators. Hence the node selection procedure just aims at identifying the
node with the largest reward gap to a given reference node.
On the other hand, if the graph has cycles, we have to rely on a more robust node
elimination procedure, akin to the one investigated in \cite{evendar2006action}
(see also the more recent \cite{abm10}).


\subsection{Large Deviations Inequalities}

In this work we use Hoeffding's maximal inequality (e.g., \cite{cesa2006}).
\begin{lemma}\label{l:hoeff}
\label{lem:Hoeffding_maximal}Let $X_{1},\ldots,X_{N}$ be independent
random variables with zero mean satisfying $a_{i}\le X_{i}\le b_{i}$
w.p. 1. Let $S_{i}=\sum_{j=1}^{i}X_{j}$. Then, \[
P\left(\max_{1\le i\le N}S_{i}>\epsilon\right)
\le\exp\left(-\frac{\epsilon^{2}}{\sum_{i=1}^{N}\left(b_{i}-a_{i}\right)^{2}}
\right).
\]
\end{lemma}
%

\section{Linear topology and sample complexity}\label{sec:Linear-Topology}
As a warm-up, we start by considering the GB setup in the case of a linear graph,
i.e., $E=\left\{ (i,i+1): 1\le i\le n-1\right\} $.
We call it the \emph{linear setup}.
The algorithm
for finding the highest node in the linear setup is presented in Algorithm
\ref{alg:Algorithm-linear}. The algorithm samples all the edges, computes for each
edge its empirical mean, and based on these statistics finds the highest edge.
Algorithm \ref{alg:Algorithm-linear} will also serve as a  subroutine for the
tree-topology discussed in Section \ref{s:tree}.
\begin{algorithm}
\begin{algorithmic}[1]
\renewcommand{\algorithmicrequire}{\textbf{Input:}}
\renewcommand{\algorithmicensure }{\textbf{Output:}}
\REQUIRE {$\epsilon>0$, $\delta>0$, line graph with edge set $E=\left\{ (i,i+1): 1\le i\le n-1\right\} $ }
\FOR{$i=1,\ldots,n-1$}
\STATE {Pull edge $(i,i+1)$ for $T^{i}$ times}
\STATE {Let $\hat{E}^{i,i+1}=\frac{1}{T^{i}}\sum_{t=1}^{T^{i}}E_{t}^{i,i+1}$ be the empirical average of edge $(i,i+1)$}
\STATE {Let $\hat{E}_{\pi_{1i}}^{1i} = \sum_{k=1}^{i-1}\hat{E}^{k-1,k} $ be the empirical average of the
composed edge $E_{\pi_{1i}}^{1i}$, where $\pi_{1i}$ is the (unique) path from 1 to $i$.}
\ENDFOR
\ENSURE {Node $k = {\rm argmax}_{i = 1,..., n} \hat{E}_{\pi_{1i}}^{1i}$.
}
\end{algorithmic}
\caption{The algorithm for the linear setup\label{alg:Algorithm-linear}}
\end{algorithm}
The following proposition gives the sample complexity of Algorithm \ref{alg:Algorithm-linear} in the case
when the edges are bounded.
\begin{proposition}\label{prop:linear_PAC_bounded}
If $-1\le E^{i,i+1}\le 1$ holds, then Algorithm \ref{alg:Algorithm-linear}
operating on a linear graph with reward gap $u$
is an $\left(\epsilon,\delta\right)$-PAC algorithm when the $T^i$
satisfy
\[
\left(\sum_{i=1}^{n-1}\frac{4}{T^{i}}\right)^{-1}\ge\frac{1}{\max\{\epsilon,u\}^{2}}\log\left(\frac{2}{\delta}\right).
\]
If $T^{i}=T$
then the sample complexity of each edge is
$
T\ge\frac{4n}{\max\{\epsilon,u\}^{2}}\log\left(\frac{2}{\delta}\right).
$\
Hence the sample complexity of the algorithm is $O\left(\frac{n^2}{\max\{\epsilon,u\}^{2}}
\log\left(\frac{1}{\delta}\right)\right).$
\end{proposition}
\begin{proof}
Let
$
\tilde{E}_{t}^{i,i+1}\triangleq\frac{(r_{i+1}-r_{i})-E_{t}^{i,i+1}}{T^{i}},\quad t = 1, ... T^{i}.
$
Each $\tilde{E}_{t}^{i,i+1}$ has zero mean with
$-2/T^{i}\le\tilde{E}_{t}^{i,i+1}\le 2/T^{i}$.
Hence
\begin{equation}\label{eq:sequence}
\tilde{E}_{1}^{1,2},\ldots,\tilde{E}_{T^{1}}^{1,2},
\tilde{E}_{1}^{2,3},\ldots,\tilde{E}_{T^{2}}^{2,3},
\ldots,
\tilde{E}_{1}^{n-1,n},\ldots,\tilde{E}_{T^{n-1}}^{n-1,n}
\end{equation}
is a sequence of $\sum_{i=1}^{n-1}T^{i}$ zero-mean and independent random
variables.
Set for brevity $\tilde{\epsilon} = \max\{\epsilon,u\}$, and
suppose, without lost of generality, that some node $j$ has the highest value.
The probability that Algorithm \ref{alg:Algorithm-linear}
fails, i.e., returns a node whose value is $\epsilon$ below the optimal value is bounded by
\begin{equation}\label{e:err}
\Pr\left(\exists i = 1,\ldots,n : \hat{E}_{\pi_{1i}}^{1i} - \hat{E}_{\pi_{1j}}^{1j}> 0 \textrm{ and }
r_i < r_j - \tilde{\epsilon} \right)~.
\end{equation}
We can write
\begin{align}
(\ref{e:err})
& \le \Pr\left(\exists i = 1,\ldots,n : \hat{E}_{\pi_{1i}}^{1i} - \hat{E}_{\pi_{1j}}^{1j} -\left(r_{i}-r_{j}\right)
> \tilde{\epsilon} \right)\notag\\
& = \Pr\left(\exists i = 1,\ldots,n : \sum_{k=1}^{i-1}\sum_{t=1}^{T^{i}}\tilde{E}_{t}^{k,k+1} > \tilde{\epsilon}\right)\notag\\
& \le \Pr\left(\exists\textrm{ partial sum in \eqref{eq:sequence} with magnitude} > \tilde{\epsilon}\right)\notag\\
 & \le2\exp\left(-\frac{\tilde{\epsilon}^{2}}{\sum_{k=1}^{n-1}\sum_{t=1}^{T^{k}}\left(2/T^{k}\right)^{2}}\right),\notag
\end{align}
%
where in the last inequality we used Lemma \ref{lem:Hoeffding_maximal}.
Requiring this probability to be bounded by $\delta$
yields the claimed inequality.
\qed
\end{proof}

The sample sizes $T^i$ in Proposition \ref{prop:linear_PAC_bounded}
encode constraints on the number of times the edges $(i,i+1)$ can be sampled.
Notice that the statement therein implies
$T_i \geq \frac{4}{\max\{\epsilon,u\}^{2}}\log\left(\frac{2}{\delta}\right)$ for all $i$,
i.e., we cannot afford in a line graph to undersample any edge. This is because
every edge in a line graph is a {\em bridge}, hence a poor estimation of any such
edge would affect the differential reward estimation throughout the graph.
In this respect, this proposition only allows for a partial tradeoff among these numbers.

\section{Tree topology and its sample complexity}\label{s:tree}
In this section we investigate PAC algorithms for finding the best
node in a tree. Let then $G = (V,E)$ be an $n$-node
tree with diameter $D$ and a set of leaves $L \subseteq V$.
Without loss of generality
we can assume that the tree is rooted at node 1 and that all edges
are directed downwards to the leaves. Algorithm \ref{alg:Algorithm-tree}
considers all possible paths from the root to the leaves and treats
each one of them as a line graph to be processed as in Algorithm
\ref{alg:Algorithm-linear}.
We have the following proposition where, for simplicity of presentation,
we do no longer differentiate among the sample sizes $T^{i,j}$ associated
with edges $(i,j)$.


%
\begin{algorithm}
\begin{algorithmic}[1]
\renewcommand{\algorithmicrequire}{\textbf{Input:}}
\renewcommand{\algorithmicensure }{\textbf{Output:}}
\REQUIRE {$\epsilon>0$, $\delta>0$, tree graph with set of leaves $L \subseteq V$}
\FOR{all leaves $k \in L$}
\STATE {Pull each edge $(i,j) \in E$ for $T$ times}
\STATE {Let $\hat{E}^{ij}=\frac{1}{T}\sum_{t=1}^{T}E_{t}^{ij}$
be the empirical average of edge $(i,j)$,
and $m_k = {\rm argmax}_{(1,i) \in \pi_{1,k}} \hat{E}_{\pi_{1i}}^{1i}$
be the maximum empirical average along path $\pi_{1k}$ (as in Algorithm \ref{alg:Algorithm-linear})}
\ENDFOR
\ENSURE {Node $m = {\rm argmax}_{k \in L} m_k$.}
\end{algorithmic}
\caption{The algorithm for the tree setup\label{alg:Algorithm-tree}}
\end{algorithm}
\begin{proposition}\label{prop:tree_PAC_bounded}
If $-1\le E^{ij}\le 1$ holds, then Algorithm \ref{alg:Algorithm-tree}
operating on a tree graph with reward gap $u$
is an $\left(\epsilon,\delta\right)$-PAC algorithm when
the sample complexity $T$ of each edge satisfies
$
T\ge\frac{4D}{\max\{\epsilon,u\}^{2}}\log\left(\frac{2|L|}{\delta}\right).
$\
Hence the sample complexity of the algorithm is
$O\left(\frac{nD}{\max\{\epsilon,u\}^{2}}\log\left(\frac{|L|}{\delta}\right)\right).$
\end{proposition}
\begin{proof}
The probability that Algorithm \ref{alg:Algorithm-tree}
returns a node whose average reward is $\epsilon$ below  the optimal one
coincides with the probability that there exists a leaf $k \in L$ such that
Algorithm \ref{alg:Algorithm-linear}
operating on the linear graph $\pi_{1,k}$ singles out a node $m_k$  whose average
reward is more than $\epsilon$ from the optimal one within $\pi_{1,k}$.
Setting
$T = \frac{4|\pi_{1,k}|}{\tilde{\epsilon}^{2}}\log\left(\frac{2|L|}{\delta}\right)$,
with $\tilde{\epsilon} = \max\{\epsilon,u\}$,
ensures that the above happens with probability at most $\delta/|L|$.
Hence each edge is sampled at most
$\frac{4D}{\tilde{\epsilon}^{2}}\log\left(\frac{2|L|}{\delta}\right)$ times and the
claim follows by a standard union bound over $L$. \qed
\end{proof}

\section{Network Sample Complexity\label{s:net}}
In this section we deal with the problem
of finding the optimal reward in a general connected and undirected
graph $G = (V,E)$, being $|V|=n$.
We describe a node elimination algorithm that works in phases,
sketch an efficient implementation and provide a sample complexity.
The following ancillary definitions will be useful.
We say that a node is a \emph{local maximum}
in a graph if all its neighboring nodes do not have higher expected reward
than the node itself. The distance between node $i$ and node $j$
is the length of the shortest path between the two nodes.
Finally, the diameter $D(G)$ of a graph $G$ is the largest distance
between any pair of nodes.

Our suggested Algorithm operates in $\log n$ phases. For notational
simplicity, it will be convenient to use subscripts to denote the
phase number.
We begin with Phase 1, where the graph $G_1 = (V_1,E_1)$
is the original graph, i.e., at the beginning all nodes are participating,
and $n_1 = |V_1| = n.$
We then find a subgraph of $G_1$, which we call
\emph{sampled graph} denoted by $G_{1}^{S}$,
that includes all the edges involved in shortest paths between
all nodes in $V_1$. We sample each edge in subgraph $G_{1}^{S}$ for $T_{1}$-times,
and compute the corresponding sample averages. Based on these averages,
we find the local maxima\footnote
{
Ties can be broken arbitrarily.
}
of $G_{1}^{S}$.

The key observation is that there can be at most $n_{1}/2$ maxima.
Denote this set of maxima by $V_{2}$. Now, define a subgraph, denoted by
$G_{2}$, whose nodes are $V_{2}$. We repeat the process of getting
a sampled graph, denoted by $G_{2}^{S}$. We sample the edges of the
sampled graph $G_{2}^{S}$ for $T_{2}$-times and define, based on its maxima, a new subgraph.
Denote the set of maxima by $V_{3}$, and the process continues until
only one node is left. We call this algorithm NNE (\emph{Network Node Elimination}),
which is similar to the action elimination procedure of \cite{evendar2006action}
(see also \cite{abm10}). The algorithm
is summarized in Algorithm \ref{alg:network}.
\begin{algorithm}[h]
\begin{algorithmic}[1]
\renewcommand{\algorithmicrequire}{\textbf{Input:}}
\renewcommand{\algorithmicensure }{\textbf{Output:}}
\REQUIRE {$\epsilon>0$, $\delta>0$, graph $G=(V,E)$, $i=1$}
\STATE {Initialize $G_1=G$, $V_1=V$}
\STATE {Compute the shortest path between all pairs of nodes of $G_1$,
and denote each path by $\pi_{ij}$. }
\STATE {Initialize the shortest path set by $SP_1 = \{\pi_{ij} | i,j \in V_1 \}$ }
\WHILE{$|V_i|>1$}
\STATE {$n_i=|V_i|$}
\STATE {Using the shortest paths in $SP_i$, find a sampled graph $G^S_i$ of $G_i$}
\STATE {$D_{i} = D(G^S_{i})$}
\STATE {Pull each edge in $G^S_i$ for $T_i$ times}
\STATE {Find the local maxima set, $V_{i+1}$, on $G^S_i$,
and get a subgraph $G_{i+1}$ that contains $V_{i+1}$}
\STATE {$SP_{i+1} = \{\pi_{ij} \in SP_i | i,j \in V_{i+1} \}$ }
\STATE{$i \leftarrow i+1$}
\ENDWHILE
\ENSURE {The remaining node}
\end{algorithmic}
\caption{The Network Node Elimination Algorithm\label{alg:network}}
\end{algorithm}
Two points should be made regarding the NNE algorithm. First, as
will be observed below, the sequence $\left\{ D(G_{i}^{S})\right\} _{i=1}^{\log n}$
of diameters is nonincreasing. Second, from the implementation viewpoint, a data-structure
maintaining all shortest paths between nodes is crucial, in order to efficiently
eliminate nodes while tracking the shortest paths between the surviving
nodes of the graph. In fact, this data structure might just be a collection of $n$
breadth-first spanning trees rooted at each node, that encode the shortest path between the root
and any other node in the graph. When node $i$ gets eliminated, we first eliminate the
spanning tree rooted at $i$, but also prune all the other spanning trees where node $i$
occurs as a leaf. If $i$ is a (non-root) internal node of another tree,
then $i$ should not be eliminated from this tree
since $i$ certainly belongs to the shortest path between another pair of surviving nodes. Note that connectivity is maintained through the process.

The following result gives a PAC bound for Algorithm \ref{alg:network} in the case
when the $E^{i,j}$ are bounded.
\begin{proposition}\label{lem:network}
Suppose that $-1\le E^{i,j}\le 1$ for every $(i,j)\in E$. Then
Algorithm \ref{alg:network} operating on a general graph $G$ with diameter $D$ and
reward gap $u$ is an $\left(\epsilon,\delta\right)$-PAC
algorithm with edge sample complexity
\begin{align}
T   \le ~\frac{\sum_{i=1}^{\log n}n_i D_i}{\left(\max\{\epsilon,u\}/\log n\right)^2}
\,\log\left(\frac{n}{\delta/\log n}\right)
\le \frac{n\,D}{\left(\max\{\epsilon,u\}/\log n \right)^2}
\,\log\left(\frac{n}{\delta/\log n}\right).\notag
\end{align}
\end{proposition}%
\begin{proof}
In each phase we have at most half the nodes of the previous
phase, i.e., $n_{i+1}\le n_{i}/2$. Therefore, the algorithm stops
after at most $\log n$ phases. Also, because we retain shortest path between
the surviving nodes, we also have $D_{i+1} \leq D_i \leq D$. At each phase, similar
to the previous sections, we make sure that it is at most $\delta/\log n$
the probability of identifying an $\epsilon/\log n$-optimal
node.
%
%
Therefore, it suffices to pull the edges in each sampled graph $G_{i}^{S}$
for $T_{i}\le\frac{n_i D_i}{\left(\max\{\epsilon,u\}/\log (n) \right)^2}
\log\left(\frac{n_i}{\delta/\log n}\right)$
times. Hence the overall sample complexity for an $\left(\epsilon,\delta\right)$-PAC
bound is at most $\sum_{j=1}^{\log n} T_j$, as claimed.
The last inequality just follows from  $n_{i+1}\le n_{i}/2$
and $D_i \leq D$ for all $i$.
%
\qed
\end{proof}
Being more general, the bound contained in Proposition \ref{lem:network} is weaker
than the ones in previous sections when specialized to line graphs or trees.
In fact, one is left wondering whether it is always convenient to
reduce the identification problem on a general graph $G$ to the
identification problem on trees by, say, extracting a suitable spanning tree
of $G$ and then invoking Algorithm \ref{alg:Algorithm-tree} on it.
The answer is actually negative, as the set of simulations reported
in the next section show.

\section{Simulations\label{s:sims}}
In this section we briefly investigate the role of the graph topology
in the sample complexity.



\begin{figure}[t]
\begin{tabular}{c@{\hspace{3.1pc}}c@{\hspace{2.5pc}}c}
\ \ \ & \figscale{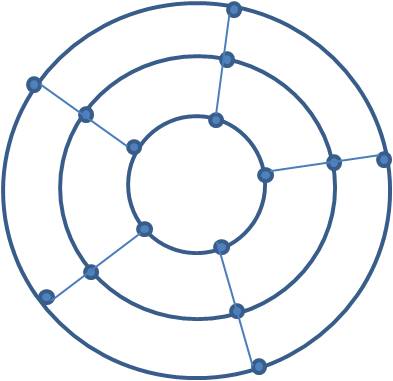}{0.5} &\figscale{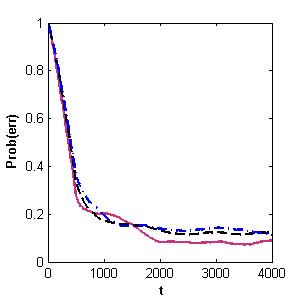}{0.5}\\
\ \ \ & (a)                            &(b)
\end{tabular}
\caption{
$(a)$ The spider-web topology.
$(b)$ Empirical error vs. time for the graph setup in (a) and spanning trees thereof.
Three algorithms are compared: NNE (red solid line),
the tree-based algorithm operating on a smallest diameter spanning tree
(black dashed line), and the tree-based algorithm operating on a
largest diameter spanning tree (blue dash-dot line).
The parameters are $n=15$ and $\epsilon=0$.
Average of 200 runs.
}
\label{fig:linear}
\end{figure}

In our simple experiment we compare Algorithm \ref{alg:Algorithm-tree}
(with two types of spanning trees) to Algorithm \ref{alg:network} over the
``spider web graph" illustrated in Figure \ref{fig:linear} (a).
This graph is made up of  15 nodes arranged in 3 concentric circles
(5 nodes each), where the circles are connected so as to resemble a spider web.
Node rewards are independently generated from the uniform distribution
on [0,1], edge rewards are just uniform in [-1,+1].
The two mentioned spanning trees are the longest diameter spanning tree (diameter 14)
and the shortest diameter spanning tree (diameter 5).
As we see from Figure \ref{fig:linear} (b), the latter tends to outperform the former.
However, both spanning tree-based algorithms are eventually outperformed by
NNE on this graph. This is because in later stages
NNE tends to handle smaller subgraphs, hence it needs only compare subsets of "good nodes".


\section{Extensions}\label{s:extensions}
We now sketch an extension of our framework to the case when the algorithm
receives contextual information in the form of feature vectors before sampling the edges.
This is intended to model a more practical setting where, say, a web system has preliminary
access to a set of user profile features.

This extension is reminiscent of the so-called {\em contextual bandit} learning
setting (e.g., \cite{lz07}), also called {\em bandits with covariates} (e.g., \cite{rz10}).
In such a setting, it is reasonable to assume that different users $\bx_s$ have
different preferences (i.e., different best nodes associated with), but also that similar
users tend to have similar preferences.
A simple learning model that accommodates the above
(and is also amenable to theoretical analysis) is to assume each node $i$ of $G$ to host a
linear function $\bu_i\,:\, \bx \rightarrow \bu_i^{\top}\bx$ where, for simplicity,
$||\bu_i|| = ||\bx|| = 1$ for all $i$ and $\bx$.
The optimal node $i^*(\bx)$ corresponding to vector $\bx$ is
$i^*(\bx) = \arg\max_{i \in V} \bu_i^{\top}\bx$. Our goal is to identify,
for the given $\bx$ at hand, an $\epsilon$-optimal node $j$ such that
$\bu_j^{\top}\bx \geq \bu_{i^*}^{\top}\bx -\epsilon$.
Again, we do not directly observe node rewards, but only the differential rewards
provided by edges.\footnote
{
For simplicity of presentation, we disregard the reward gap here.
}
When we operate on input $\bx$ and pull edge $(i,j)$,
we receive an independent observation of random variable
$E^{ij}(\bx)$ such that
$\mathbb{E}[E^{ij}(\bx)] = \bu_j^{\top}\bx- \bu_i^{\top}\bx$.

Learning proceeds in a sequence of {\em stages} $s=1, \ldots, S$, each stage
being in turn a sequence of time steps corresponding to the edge pulls taking
place in that stage. In Stage 1 the algorithm gets input $\bx_1$, is allowed
to pull (several times) the graph edges $E^{ij}(\bx_1)$, and is required to
output an $\epsilon$-optimal node for $\bx_1$. Let $T(\bx_1)$ be the sample complexity
of this stage. In Stage 2, we retain the information gathered in Stage 1,
receive a new vector $\bx_2$ (possibly close to $\bx_1$) and repeat the same kind of
inference, with sample complexity $T(\bx_2)$. The game continues until $S$ stages
have been completed.

For any given sequence $\bx_1$, $\bx_2$, $\ldots$, $\bx_S$, one expects the cumulative
sample size
\(
\sum_{s=1}^S T(\bx_s)
\)
to grow {\em less than linearly} in $S$. In other words, the additional effort the
algorithm makes
in the identification problem diminishes with time, as more and more users are interacting
with the system, especially when these users are similar to each other, or even occur
more than once in the sequence $\bx_1$, $\bx_2$, $\ldots$, $\bx_S$.
In fact, we can prove stronger results of the following kind. Notice that the bound
does not depend on the number $S$ of stages, but only on the dimension of the input space.\footnote
{
A slightly different statement holds in the case when the input dimension is infinite.
This statement quantifies the cumulative sample size w.r.t.
the amount to which the vectors $\bx_1$, $\bx_2$, $\ldots$, $\bx_S$ are close to each other.
Details are omitted due to lack of space.
}
\begin{proposition}
Under the above assumptions, if $G = (V,E)$ is a connected and undirected graph, with $n$ nodes and
diameter $D$, and $\bx_1$, $\bx_2$, $\ldots$, $\bx_S \in R^d$  is any sequence of unit-norm feature vectors,
then with probability at least $1-\delta$ a version of
the NNE algorithm exists which outputs at each stage $s$ an $\epsilon$-optimal
node for $\bx_s$, and achieves the following cumulative sample size
\[
\sum_{s=1}^S T(\bx_s) = O\left( B\,\log^2 B\right),
\]
where $B = \frac{n\,D}{\left(\epsilon/\log n\right)^2}\,\log\left(\frac{n}{\delta/\log n}\right)\,d^2.$
%
\end{proposition}
\begin{proof}[Sketch]
The algorithm achieving this bound combines linear-regression-like estimators with NNE. In particular,
every edge of $G$ maintains a linear estimator ${\hat \bu^{ij}}$ intended to approximate the difference
$\bu_j-\bu_i$ over both stages and sampling times within each stage.
At stage $s$ and sampling time $t$ within stage $s$,
the vector ${\hat \bu^{ij}_{s,t}}$ suitably stores all past feature vectors $\bx_1, \ldots, \bx_{s}$
observed so far, along with the corresponding edge reward observations.
By using tools from ridge regression in adversarial settings (see, e.g., \cite{dgs10}),
one can show high-probability approximation results of the form
\begin{align}
({\hat \bu^{ij}_{s,t}} \,^{\top}\bx - (\bu_j-\bu_i)^{\top}\bx)^2
\leq \bx^{\top}A_{s,t}^{-1}\bx\,\left(d\,\log \Sigma_{s,t} + \log\frac{1}{\delta}\right), \label{e:approx}
\end{align}
being $\Sigma_{s,t} = \sum_{k \leq s-1} T(\bx_k) + t$, and
$A_{s,t}$ the matrix
\[
A_{s,t} = I + \sum_{k \leq s-1} T(\bx_k)\bx_k\bx_k^{\top} + t\,\bx_s\bx_s^{\top}~.
\]
In stage $s$, NNE is able to output an $\epsilon$-optimal node for input $\bx_s$
as soon as the RHS of (\ref{e:approx}) is as small as $c\epsilon^2$, for a
suitable constant $c$
depending on the current graph topology NNE is operating on.
Then the key observation is that in stage $s$ the number of times we sample an edge $(i,j)$
such that the above is false cannot be larger than
\[
\frac{1}{c\epsilon^2}\,\log \frac{|A_{s,T(\bx_s)}|}{|A_{s,0}|}\left(d\,\log \Sigma_{s,T(\bx_s)} + \log\frac{1}{\delta}\right),
\]
where $|\cdot|$ is the determinant of the matrix at argument.
This follows from standard inequalities of the form
$\sum_{t=1}^{T(\bx_s)} \bx_s^{\top}A_{s,t}^{-1}\bx_s \leq \log \frac{|A_{s,T(\bx_s)}|}{|A_{s,0}|}$.\qed
\end{proof}

\section{Discussion}\label{s:discussion}


This paper falls in the research thread of
analyzing online decision problems where
the information that is obtained is comparative between arms. We analyzed
a simple setup where the structure of comparisons is provided by a given graph
which, unlike previous works on this subject \cite{yue2011,yj11}, lead
us focus on the notion of
finding an $\epsilon$-optimal arm with high probability. We then described
an extension to the important contextual setup.
There are several issues that call for further research that we outline below.


First, we only addressed the exploratory bandit problem. It would be interesting to consider
the regret minimization version of the problem. While naively one can think of it as a problem
with an arm per edge of the graph, this may not be a very
effective model because the number of arms may go as $n^2$ but the number of parameters grows
like $n$. On top of this, definining a meaningful notion of regret may not be trivial
(see the discussion in the introductory section).
Second, we only considered graphs as opposed to hypergraphs. Considering comparisons of more
than two nodes raises interesting modeling issues and well as computational issues.
Third, we assumed that all samples are equivalent in the sense that all the pairs we can
compare have the same cost. This is not a realistic assumption in many applications.
An approach akin to budgeted learning \cite{Madani04thebudgeted} would be interesting here.
Fourth, we focused on upper bounds and constructive algorithms. Obtaining lower bounds
that depend on the network topology would be interesting. The upper bounds we have provided
are certainly loose for the case of a general network.
Furthermore, more refined upper bounds are likely to exist which take into account the distance
on the graph between the good nodes (e.g., between the best and the second best ones).
In any event, the algorithms we developed for the network case are certainly not optimal.
There is room for improvement by reusing information better and by adaptively selecting which
portions of the network to focus on. This is especially interesting
under smoothness assumptions on the expected rewards. Relevant references in the MAB setting
to start off with include \cite{a07,ksu08,bmss08}.


\end{document}